\documentclass{article}
\pdfoutput=1

\usepackage[protrusion=true,tracking=true,kerning=true,expansion,final]{microtype}
\usepackage{graphicx}
\usepackage{subfigure}
\usepackage{booktabs} 

\usepackage[pagebackref]{hyperref}

\usepackage[accepted]{sysml2019}

\usepackage{amsfonts}
\usepackage{amssymb}
\usepackage{amsmath}
\usepackage{graphicx}
\usepackage{multirow}
\usepackage{makecell}
\usepackage{flafter} 
\usepackage{physics} 
\usepackage{esint} 
\usepackage{docmute} 
\usepackage{soul}

\usepackage{xcolor}

\definecolor{color_iqn}{RGB}{60,180,75}
\definecolor{color_xnor}{RGB}{255,225,25}
\definecolor{color_appr}{RGB}{0,128,128}
\definecolor{color_qsm}{RGB}{245, 130, 48}
\definecolor{color_dist}{RGB}{145, 30, 180}
\definecolor{color_comp}{RGB}{128, 128, 0}
\definecolor{color_qnn}{RGB}{0, 0, 128}
\definecolor{color_mlq}{RGB}{175, 50, 125} 
\definecolor{color_ours}{RGB}{255,0,0}
\sysmltitlerunning{UNIQ: Uniform Noise Injection for Quantization}

\begin{document}

\twocolumn[
	\sysmltitle{UNIQ: Uniform Noise Injection for\\Non-Uniform Quantization of Neural Networks}

	\sysmlsetsymbol{equal}{*}

	\begin{sysmlauthorlist}
		\sysmlauthor{Chaim Baskin}{equal,tech}
		\sysmlauthor{Eli Schwartz}{equal,ta}
		\sysmlauthor{Evgenii Zheltonozhskii}{tech}
		\sysmlauthor{Natan Liss}{tech}
		\sysmlauthor{Raja Giryes}{ta}
		\sysmlauthor{Alex Bronstein}{tech}
		\sysmlauthor{Avi Mendelson}{tech}
	\end{sysmlauthorlist}

	\sysmlaffiliation{tech}{Department of Computer Science, Technion, Haifa, Israel}
	\sysmlaffiliation{ta}{School of Electrical Engineering, Tel-Aviv University, Tel-Aviv, Israel}

	\sysmlcorrespondingauthor{Evgenii Zheltonozhskii}{evgeniizh@campus.technion.ac.il}
	\sysmlcorrespondingauthor{Eli Schwartz}{eliyahus@mail.tau.ac.il}

	\sysmlkeywords{Machine Learning, SysML, Neural Networks, Quantization, Efficient Inference, Deep Learning}

	\vskip 0.3in

	\begin{abstract}
		We present a novel method for neural network quantization that emulates a non-uniform $k$-quantile quantizer, which adapts to the distribution of the quantized parameters. 
		Our approach provides a novel alternative to the existing uniform quantization techniques for neural networks. 
		We suggest to compare the results as a function of the bit-operations (BOPS) performed, assuming a look-up table availability for the non-uniform case. 
		In this setup, we show the advantages of our strategy in the low computational budget regime. 
		While the proposed solution is harder to implement in hardware, we believe it sets a basis for new alternatives to neural networks quantization.
	\end{abstract}
]



\printAffiliationsAndNotice{\sysmlEqualContribution} 

\begin{figure*}[t]
	\begin{center}
		\centerline{\includegraphics[width=0.9\linewidth]{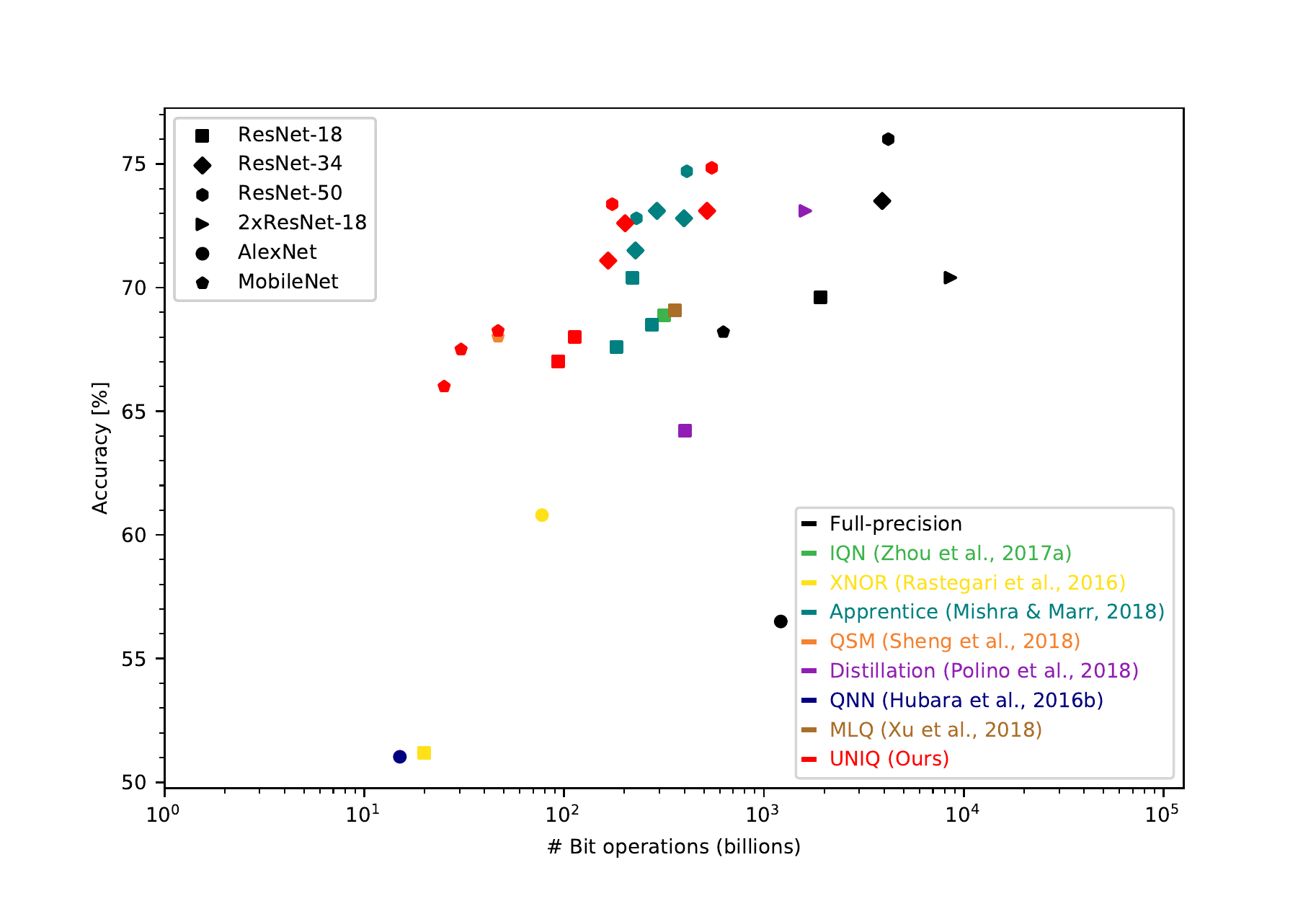}}
		\caption{
			{\bf Performance vs.\ complexity of different quantized neural networks.} Performance is measured as top-1 accuracy on ImageNet; complexity is estimated in number of bit operations. Network architectures are denotes by different marker shapes; quantization methods are marked in different colors. Multiple instances of the same method (color) and architecture (shape) represent different level of quantization. Notice that our optimized networks are the most efficient ones among all the ones that achieves accuracy 73.4\% or lower and are the most accurate ones among all the ones that consume $400$ GBOPs or lower.
			The figure should be viewed in color.
		}
		\label{fig_accuracy_vs_bops}
	\end{center}
	\vskip -0.2in
\end{figure*}

\begin{table*}[ht]
	\caption{Complexity-accuracy tradeoff of various DNN architectures quantized  using different techniques. Complexity is reported in number of bit operations as explained in the text. Number of bits is reported as (weights,activations) and model size is calculated as the sum of parameters sizes in bits. Accuracy is top-1 accuracy on ImageNet. For each DNN architecture, rows are sorted in increasing order complexity. Note that for the same architecture with same bitwidths UNIQ may have a different complexity and model size due to fact that, unlike the others, we also quantize the first and the last layers.  Compared methods:  \textcolor{color_xnor}{XNOR} \cite{rastegari2016xnor}, \textcolor{color_qnn}{QNN} \cite{hubara2016quantized}, \textcolor{color_iqn}{IQN} \cite{zhou2017incremental}, \textcolor{color_appr}{Apprentice} \cite{mishra2017apprentice},  \textcolor{color_dist}{Distillation} \cite{polino2018model}, \textcolor{color_qsm}{QSM} \cite{DBLP:journals/corr/abs-1803-08607}, \textcolor{color_mlq}{MLQ} \cite{AAAI1816479}. }
	\label{tab_resnet_imagenet_comparison}
	\begin{center}
		\begin{small}
			\begin{tabular}{llcccc}
				\toprule
				                  &
				                  &
				\bf{Bits}         &
				\bf{Model size}   &
				\bf{Complexity}   &
				\bf{Accuracy}                                                                                \\

				\bf{Architecture} &
				\bf{Method}       &
				\bf{(w,a)}        &
				\bf{[Mbit]}       &
				\bf{[GBOPs]}      &
				\bf{(\% top-1)}                                                                              \\

				\midrule
				AlexNet           & \textcolor{color_qnn}{QNN}           & 1,2   & 15.59  & $15.1 $  & 51.03 \\
				AlexNet           & \textcolor{color_xnor}{XNOR}         & 1,32  & 15.6   & $77.5 $  & 60.10 \\
				AlexNet           & Baseline                             & 32,32 & 498.96 & $1210 $  & 56.50 \\
				\midrule
				MobileNet         & \textcolor{color_ours}{UNIQ} (Ours)  & 4,8   & 16.8   & $25.1 $  & 66.00 \\
				MobileNet         & \textcolor{color_ours}{UNIQ} (Ours)  & 5,8   & 20.8   & $30.5 $  & 67.50 \\
				MobileNet         & \textcolor{color_ours}{UNIQ} (Ours)  & 8,8   & 33.6   & $46.7$   & 68.25 \\
				MobileNet         & \textcolor{color_qsm}{QSM}           & 8,8   & 33.6   & $46.7 $  & 68.01 \\
				MobileNet         & Baseline                             & 32,32 & 135.2  & $626$    & 68.20 \\
				\midrule
				ResNet-18         & \textcolor{color_xnor}{XNOR}         & 1,1   & 4      & $19.9 $  & 51.20 \\
				ResNet-18         & \textcolor{color_ours}{UNIQ} (Ours)  & 4,8   & 46.4   & $93.2$   & 67.02 \\
				ResNet-18         & \textcolor{color_ours}{UNIQ} (Ours)  & 5,8   & 58.4   & $113$    & 68.00 \\
				ResNet-18         & \textcolor{color_appr}{Apprentice}   & 2,8   & 39.2   & $183 $   & 67.6  \\
				ResNet-18         & \textcolor{color_appr}{Apprentice}   & 4,8   & 61.6   & $220 $   & 70.40 \\
				ResNet-18         & \textcolor{color_appr}{Apprentice}   & 2,32  & 39.2   & $275 $   & 68.50 \\
				ResNet-18         & \textcolor{color_iqn}{IQN}           & 5,32  & 72.8   & $359 $   & 68.89 \\
				ResNet-18         & \textcolor{color_mlq}{MLQ}           & 5,32  & 58.4   & $359 $   & 69.09 \\
				ResNet-18         & \textcolor{color_dist}{Distillation} & 4,32  & 61.6   & $403 $   & 64.20 \\
				ResNet-18         & Baseline                             & 32,32 & 374.4  & $1920  $ & 69.60 \\
				\midrule
				ResNet-34         & \textcolor{color_ours}{UNIQ} (Ours)  & 4,8   & 86.4   & $166$    & 71.09 \\
				ResNet-34         & \textcolor{color_ours}{UNIQ} (Ours)  & 5,8   & 108.8  & $202 $   & 72.60 \\
				ResNet-34         & \textcolor{color_appr}{Apprentice}   & 2,8   & 59.2   & $227 $   & 71.5  \\
				ResNet-34         & \textcolor{color_appr}{Apprentice}   & 4,8   & 101.6  & $291  $  & 73.1  \\
				ResNet-34         & \textcolor{color_appr}{Apprentice}   & 2,32  & 59.2   & $398  $  & 72.8  \\
				ResNet-34         & \textcolor{color_ours}{UNIQ} (Ours)  & 4,32  & 86.4   & $519$    & 73.1  \\
				ResNet-34         & Baseline                             & 32,32 & 697.6  & $3930  $ & 73.4  \\
				\midrule
				ResNet-50         & \textcolor{color_ours}{UNIQ} (Ours)  & 4,8   & 102.4  & $174 $   & 73.37 \\
				ResNet-50         & \textcolor{color_appr}{Apprentice}   & 2,8   & 112.8  & $230  $  & 72.8  \\
				ResNet-50         & \textcolor{color_appr}{Apprentice}   & 4,8   & 160    & $301$    & 74.7  \\
				ResNet-50         & \textcolor{color_appr}{Apprentice}   & 2,32  & 112.8  & $411 $   & 74.7  \\
				ResNet-50         & \textcolor{color_ours}{UNIQ} (Ours)  & 4,32  & 102.4  & $548$    & 74.84 \\
				ResNet-50         & Baseline                             & 32,32 & 817.6  & $4190  $ & 76.02 \\

				\bottomrule
			\end{tabular}
		\end{small}
	\end{center}
	\vskip -0.1in
\end{table*}

\section{Introduction}
Deep neural networks are widely used in many fields today including computer vision,  signal processing, computational imaging, image processing, and speech and language processing \cite{6296526,Lai:2015:RCN:2886521.2886636,DBLP:journals/corr/ChenPK0Y16}. 
Yet, a major drawback of these deep learning models is their storage and computational cost. Typical deep networks comprise millions of parameters and require billions of multiply-accumulate (MAC) operations. In many cases, this cost renders them infeasible for running on mobile devices with limited resources. While some applications allow moving the inference to the cloud, such an architecture still incurs significant bandwidth and latency limitations.

A recent trend in research focuses on developing lighter deep models, both in their memory footprint and computational complexity. The main focus of most of these works, including this one, is on alleviating complexity at inference time rather than simplifying training. While training a deep model requires even more resources and time, it is usually done offline with large computational resources.

One way of reducing computational cost is quantization of weights and activations. Quantization of weights also reduces storage size and memory access. The bit widths of activations and weights affect linearly the amount of required hardware logic; reducing both bitwidths by a factor of two reduces the amount of hardware by roughly a factor of four.
(A more accurate analysis of the effects of quantization is presented in Section \ref{BOPS}). Quantization allows to fit bigger networks into a device, which is especially critical for embedded devices and custom hardware. On the other hand, activations are passed between layers, and thus, if different layers are processed separately, activation size reduction reduces communication overheads.

Deep neural networks are usually trained and operate with both the weights and the activations represented in single-precision ($32$-bit) floating point. A straightforward uniform quantization of the pre-trained weights to $16$-bit fixed point representation has been shown to have a negligible effect on the model accuracy \cite{gupta2015deep}. In the majority of the applications, further reduction of precision quickly degrades performance; hence, nontrivial techniques are required to carry it out.

\textbf{Contribution.} Most previous quantization solution for neural networks assume a uniform distribution of the weights, which is non-optimal due to the fact these rather have bell-shaped distributions \cite{han2015deep}.
To facilitate this deficiency, we propose a $k$-quantile quantization method with balanced (equal probability mass) bins, which is particularly suitable for neural networks, where outliers or long tails of the bell curve are less important.
We also show a simple and efficient way of reformulating this quantizer using a ``uniformization trick''. 

In addition, since most existing solutions for quantization struggle with training networks with quantized weights, we introduce a novel method for training a network that performs well with quantized values. This is achieved by adding noise (at training time) to emulate the quantization noise introduced at inference time. The uniformization trick renders exactly the injection of uniform noise and alleviates the need to draw the noise from complicated and bin-dependent distributions. While we limit our attention to the $k$-quantile quantization, the proposed scheme can work with any threshold configuration while still keeping the advantage of uniformly distributed noise in every bin.

Our proposed solution is denoted UNIQ (uniform noise injection for non-uniform qunatization).
Its better noise modeling extends the uniform noise injection strategy in \cite{Baskin18NICE} that relies on uniform quantization and thus get inaccurate quantization noise modeling in the low bit regime. 
The accurate noise model proposed here leads to an advantage in the low bit regime. Measuring the complexity of each method by bit operations (BOPS), our UNIQ strategy achieves improved performance in the accuracy vs. complexity tradeoff
compared to other leading methods such as distillation \cite{polino2018model} and XNOR \cite{rastegari2016xnor}. In addition, our method performs well on smaller models targeting mobile devices, represented in the paper by MobileNet. Those network are known to be harder to quantize, due to lower redundancy in parameters. 

One may argue that due to the non-uniform quantization, it is harder to implement our proposed solution in hardware. Yet, this can be alleviated by using look-up tables for calculating the different operations between the quantized values. Nonetheless, we believe that it is important to explore new solutions for network quantizations beyond the standard uniform strategy and this work 
is doing such a step.

\section{Related work}

Previous studies have investigated quantizing the network weights and
the activations to as low as
$1$- or $2$-bit representation \cite{rastegari2016xnor,hubara2016quantized,zhou2016dorefa,dong2017learning,mishra2017wrpn}. Such extreme reduction of the range of
parameter values
greatly affects accuracy. Recent works proposed to use a wider network, i.e., one with more filters per layer, to mitigate the accuracy loss 
\cite{zhu2016trained,polino2018model}. In some approaches, e.g., \citet{zhu2016trained} and \citet{zhou2017incremental}, a learnable linear scaling layer is added after each quantized layer to improve expressiveness.

A general approach to learning a quantized model adopted in recent papers  \cite{hubara2016binarized,hubara2016quantized,zhou2016dorefa,rastegari2016xnor,cai2017deep} is to perform the forward pass using the quantized values, while keeping another set of full-precision values for the backward pass updates. In this case, the backward pass is still 
differentiable, while the forward pass is quantized. In the aforementioned papers, a deterministic or stochastic function is used at training for weight and activation quantization.
Another approach introduced by \citet{mishra2017apprentice} and \citet{polino2018model} is based on a teacher-student setup for knowledge distillation of a full precision (and usually larger) teacher model to a quantized student model. This method allows training highly accurate quantized models, but
requires training of an additional, larger network.

Most previous studies have used uniform quantization (i.e., all quantization bins are equally-sized), which is attractive due to its simplicity. However, unless the values are uniformly distributed, uniform quantization is not optimal.

Unfortunately, neural network weights are not uniform but rather have bell-shaped distributions \cite{han2015deep,Anderson2018High} as we also show
in the supplementary material.

Non-uniform quantization is utilized by \citet{han2015deep}, where the authors replace the weight values with indexes pointing to a finite codebook of shared values. \citet{Ullrich2017soft} propose to use clustering of weights as a way of quantization. They fit a Gaussian prior model to the weights and use the cluster centroids as the quantization codebook.
\citet{AAAI1816479} also use such an approach but in addition embody incremental quantization both on network and layer levels. Other examples of methods that take data distribution into account are the Bayesian quantization \cite{Louizos2017Bayesian} and the value-aware quantization \cite{park2018value}. In \citet{Louizos2017Bayesian}, sparsity priors are imposed on the network weights, providing information as to the bits to be allocated per each layer. In \citet{Molchanov2017Variational}, variational dropout, which also relies on the Bayesian framework, is proposed to prune weights in the network.
\citet{park2018value} propose to leave a small part of weights and activations (with higher values) in full-precision, while using uniform quantization for the rest.

Another approach adopted by \citet{zhou2017incremental} learns the quantizer thresholds by iteratively grouping close values and re-training the weights. \citet{lee2017lognet} utilize a logarithmic scale quantization for an approximation of the $\ell_2$-optimal Lloyd quantizer \cite{lloyd1982least}.
\citet{cai2017deep} proposed to optimize expectation of $\ell_2$ error of quantization function to reduce the quantization error. Normally distributed weights and half-normally distributed activations were assumed, that enables using a pre-calculated $k$-means quantizer. In \citet{zhou2017balanced} balanced bins are used, so each bin has the same number of samples. In some sense, this idea is the closest to our approach; yet, while \citet{zhou2017balanced} force the values to have an approximately uniform distribution, we pose no such constraint. Also, since calculating percentiles is expensive in this setting, \citet{zhou2017balanced} estimate them with means, while our method allows using the actual percentiles as detailed in the sequel.

\section{Non-uniform quantization by uniform noise injection}
To present our uniform noise injection quantization (UNIQ) method
for training a neural network amenable to operation in low-precision arithmetic, we start by outlining several common quantization schemes and discussing their suitability for deep neural networks. Then, we suggest a training procedure where during training time uniform random additive noise is injected into weights simulating the quantization error. The scheme aims at improving the quantized network performance at inference time, when regular deterministic quantization is used.

\subsection{Quantization}

Let $X$ be a random variable drawn from some distribution described by the probability density function $f_X$. Let
$T=\left\{t_i\right\}$ with $t_0=-\infty$, $t_{k}=\infty$ and $t_{i-1}<t_i$
be a set of thresholds partitioning the real line into $k$ disjoint intervals (bins) $\{[t_{i-1}, t_{i}] \}_{i=1}^{k}$, and let $Q=\{ q_i \}_{i=1}^k$ be a set of $k$ representation levels. A quantizer $\mathcal{Q}_{T,Q}$ is a function mapping each bin $[t_{i-1}, t_{i}]$ to the corresponding representation level $q_i$. We denote the quantization error by $\mathcal{E} = X - \mathcal{Q}_{T,Q}(X)$. The effect of quantization can be modeled as the addition of random noise to $X$; the noise added to the $i$-th bin admits the conditional distribution $(X - q_i) | X \in [t_{i-1}, t_{i}] $.

Most papers on neural network quantization focus on the \emph{uniform quantizer}, which has a constant bin width $t_i - t_{i-1} = \Delta$ and $q_i = (t_{i-1}+t_i)/2$, it is known to be optimal (in the sense of the mean squared error $\mathbb{E} \mathcal{E}^2$, where the expectation is taken with respect to the density $f_X$)
in the case of uniform distribution. Yet, since $X$ in neural networks is not uniform but rather bell shaped \cite{han2015deep}, in the general case the optimal choice, in the $\ell_2$ sense, is the \emph{$k$-means quantizer}. Its name follows 
the property that each representation level $q_i$ coincides with the $i$-th bin centroid (mean w.r.t. $f_X$). While finding the optimal $k$-means quantizer is known to be an NP-hard problem, heuristic procedures such as the Lloyd-Max algorithm \cite{lloyd1982least} usually produce a good approximation. The $k$-means quantizer coincides with the uniform quantizer when $X$ is uniformly distributed.

While being a popular choice in signal processing, the $k$-means quantizer encounters severe obstacles in our problem of neural network quantization. Firstly, the Lloyd-Max algorithm has a prohibitively high complexity to be used in every forward pass. Secondly, it is not easily amenable to our scheme of modeling quantization as the addition of random noise, as the noise distribution at every bin is complex and varies with the change of the quantization thresholds. Finally, our experiments shown in Section \ref{quantization_analisys} in the sequel suggest that the use of the $\ell_2$ criterion for quantization of deep neural classifier does not produce the best classification results. The weights in such networks typically assume a bell-shaped distribution with tails exerting a great effect on the mean squared error, yet having little impact on the classification accuracy.

Based on empirical observations, we conjecture that the distribution tails, which $k$-means is very sensitive to, are not essential for good model performance at least in classification tasks. As an alternative to $k$-means, we propose the \emph{$k$-quantile quantizer} characterized by the equiprobable bins property, that is, $\mathbb{P}( X \in [t_{i-1}, t_{i}]) = 1/k$. The property is realized by setting $t_i = F_X^{-1}(i/k)$, where $F_X$
denotes the cumulative distribution function of $X$ and, accordingly, its inverse $F_X^{-1}$ denotes the quantile function.
The representation level of the $i$-th bin is set to the bin median, $q_i = \mathrm{med}\{ X | X \in [t_{i-1}, t_{i}]\}$. It can be shown that in the case of a uniform $X$, the $k$-quantile quantizer coincides with the $k$-level uniform quantizer.

The cumulative distribution $F_X$ and the quantile function $F_X^{-1}$ can be estimated empirically from the distribution of weights, and updated in every forward pass. Alternatively, one can rely on the empirical observation that the $\ell_2$-regularized weights of each layer tend to
follow an approximately normal distribution \cite{blundell2015weight}. To confirm that this is the case for the networks used in the paper, we analyzed the distribution of weights. An example of a layer-wise distribution of weights is shown in the supplementary material. Relying on this observation, we can estimate $\mu$ and $\sigma$ per each layer and use the CDF of the normal distribution (and its inverse, the normal quantile function).

Using the fact that applying a distribution function $F_X$ of $X$ to itself results in uniform distribution allows
an alternative construction of the $k$-quantile quantizer.
We apply the transformation $U=F_X(X)$ to the input converting it into a uniform random variable on the interval $[0,1]$. Then, a \emph{uniform} $k$-level quantizer (coinciding with the $k$-quantile quantizer) is applied to $U$ producing $\hat{U} = \mathcal{Q}_{\mathrm{uni}}(U)$; the result is transformed back into $\hat{X} = F_X^{-1}(\hat{U})$ using the inverse transformation. We refer to this procedure as to the \emph{uniformization trick}. Its importance will become evident in the next section.

\subsection{Training quantized neural networks by uniform noise injection}
\label{sec:uniq}

The lack of continuity, let alone smoothness, of the quantization operator renders impractical its use in the backward pass. As an alternative, at training we replace the quantizer by the injection of random additive noise. This scheme suggests that instead of using the quantized value $\hat{w} = \mathcal{Q}_{T,Q}(w)$ of a weight $w$ in the forward pass, $\hat{w} = w + e$ is used with $e$ drawn from the conditional distribution of $(W - q_i) | W \in [t_{i-1}, t_{i}] $ described by the density
$$
	f_{E}(e) = \frac{f_W(e+q_i)}{\int_{t_{i-1}}^{t_i} f_W(w) \dd{w} }
$$
defined for $e \in [t_{i-1}-q_i, t_{i}-q_i]$ and vanishing elsewhere.
The bin $i$ to which $w$ belongs is established according to its value and is fixed during the backward pass. Quantization of the network activations is performed in the same manner.

The fact that the parameters do not directly undergo quantization keeps the model differentiable. In addition, gradient updates in the backward pass have an immediate impact on the forward pass, in contrast to the directly quantized model, where small updates often leave the parameter in the same bin, leaving it effectively unchanged.

While it is customary to model the quantization error as noise with uniform distribution $f_{E}$  \cite{gray1998quantization}, this approximation breaks in the extremely low precision regimes (small number of quantization levels $k$) considered here. Hence, the injected noise has to be drawn from a potentially non-uniform distribution which furthermore changes as the network parameters and the quantizer thresholds are updated.

To overcome this difficulty, we resort to the uniformization trick outlined in the previous section. Instead of the $k$-quantile quantizer $w' = \mathcal{Q}(w)$, we apply the equivalent uniform quantizer to the uniformized variable,
$\hat{w} = F_W^{-1}(\mathcal{Q}_{\mathrm{uni}}(F_W(w)))$. The effect of the quantizer can be again modeled using noise injection,
$\hat{w} = F_W^{-1}(F_W(w) + e)$,
with the cardinal difference than now the noise $e$ is \emph{uniformly distributed} on the interval $[-\frac{1}{2k},\frac{1}{2k}]$ (estimating quantization error distribution).

Usually, quantization of neural networks is either used for training a model from scratch or applied post-training as a fine-tuning stage. Our method, as will be demonstrated in our experiments, works well in both cases. Our practice shows that best results are obtained when the learning rate is reduced as the noise is added; we explain this by the need to compensate for noisier gradients.

\subsection{Gradual quantization}
\label{gradual_quantization_method}
The described method works well ``as is'' for small- to medium-sized neural networks. For deeper networks, the basic method does not perform as well, most likely due to
errors arising when applying a long sequence of operations where more noise is added at each step. We found that applying the scheme gradually to small groups of layers works better in deep networks.
In order to perform gradual quantization, we split the network into $N$ blocks $\{B_1,...,B_N\}$, each containing about same number of consecutive layers. We also split our budget of training epochs into $N$ stages. At the $i$-th stage, we quantize and freeze the parameters in blocks $\{B_1,...,B_{i-1}\}$, and inject noise into the parameters of $B_{i}$. For the rest of the blocks  $\{B_{i+1},...,B_N\}$ neither noise nor quantization is applied.

This approach is similar to one proposed by \citet{AAAI1816479}. This way, the number of parameters into which the noise is injected simultaneously is reduced, which allows better convergence. The deeper blocks gradually adapt to the quantization error of previous ones and thus tend to converge relatively fast when the noise is injected into them. For fine-tuning a pre-trained model, we use the same scheme, applying a single epoch per stage.
An empirical analysis of the effect of the different number of stages is presented in the supplementary material. 
This process can be performed iteratively,
restarting from the beginning after the last layer has been trained. Since this allows earlier blocks to adapt to the changed values of the following ones, the iterative process yields an additional increase in accuracy. Two iterations were performed in the reported experiments.

\subsection{Activations quantization}
While we mostly deal with weights quantization, activations quantization is also beneficial in lowering the arithmetic complexity and can help decreasing the communication overhead in the distributed model case.
We observed that a by-product of training with noisy weights is that the model becomes less sensitive to a certain level of quantization of the activations. When training with the gradual process described above, activations of the fixed layers are quantized as they would be at inference time. The effect of quantized activations on classification accuracy is evaluated in the sequel.


\section{Experimental evaluation}
We performed an extensive performance analysis of the UNIQ scheme compared to the current state-of-the-art methods for neural network quantization. The basis for comparison is the accuracy vs.\ the total number of bit operations in visual classification tasks on the ImageNet-1K \cite{ILSVRC15} dataset. CIFAR-10 dataset \cite{krizhevsky2009learning} is used to evaluate different design choices made, while the main evaluation is performed on ImageNet-1K.

MobileNet \cite{howard2017mobilenets} and ResNet-18/34/50 \cite{he2016deep} architectures are used as the baseline for quantization. MobileNet is chosen as a representative of lighter models, which are more suited for a limited hardware setting where quantization is also most likely to be used. ResNet-18/34/50  is chosen due to its near state-of-the-art performance and popularity, which makes it an excellent reference for comparison.

We adopted the number of bit operations (BOPs) metric to quantify the network arithmetic complexity. This metric is particularly informative about the performance of mixed-precision arithmetic especially in hardware implementations on FPGAs and ASICs.

\paragraph{Training details}
\label{train_procedure}

For quantizing a pre-trained model, we train with SGD for a number of epochs equal to the number of trainable layers (convolution and fully connected). We also follow the gradual process described above with the number of stages set to the number of trainable layers. The learning rate is $10^{-4}$, momentum $0.9$ and weight decay $10^{-4}$. In all experiments, unless otherwise stated, we fine-tuned a pre-trained model taken from an open-sourced repository of pre-trained models.
\footnote{\href{https://github.com/Cadene/pretrained-models.pytorch}{https://github.com/Cadene/pretrained-models.pytorch}}

\subsection{Performance of quantized networks on ImageNet}
\label{imagenet_results}

Table \ref{tab_resnet_imagenet_comparison} compares the ResNet and MobileNet performance with weights and activations quantized to several levels using UNIQ and other leading approaches reported in the literature. For baseline, we use a full-precision model with $32$ bit weights and activations.
Note that the common practice of not quantizing first and last layers significantly increases the network complexity in BOPs. Since our model does quantize all the layers, including the first one, which often appears to be the most computationally-intensive, it achieve lower complexity requirements with a higher bitwidth. Note also the diminishing impact of the weights bitwidth on the BOP complexity as explained hereafter.

We found UNIQ to perform well also with the smaller MobileNet architecture.
This is in contrast to most of the previous methods that resort to larger models and  doubling the number of filters
\cite{polino2018model}, thus quadrupling the number of parameters, e.g., from $11$ to $44$ million for ResNet-18.

\subsection{Accuracy vs.\ complexity trade-off}
\label{BOPS}

Since custom precision data types are used for the network weights and activations, the number of MAC operations is not an appropriate metric to describe the computational complexity of the model. Therefore, we use the BOPs metric quantifying the number of bit operations. Given the bitwidth of two operands, it is possible to approximate the number of bit operations required for a basic arithmetic operation such as addition and multiplication.
The proposed metric is useful when the inference is performed on custom hardware like FPGAs or ASICs. Both are a natural choice for quantized networks, due to the use of lookup tables (LUTs) and dedicated MAC (or more general DSP) units, which are efficient with custom data types.

An important phenomenon that can be observed in Table \ref{tab_resnet_imagenet_comparison} is the non-linear relation between the number of activation and weight bits and the resulting network complexity in BOPs.
To quantify this effect, let us consider a single convolutional layer with $b_\mathrm{w}$-bit weights and $b_\mathrm{a}$-bit activations containing $n$ input channels, $m$ output channels, and $k \times k$ filters. The maximum value of a single output is about $2^{b_\mathrm{a}+b_\mathrm{w}} n k^2$, which sets the accumulator width in the MAC operations to $b_\mathrm{o} = b_\mathrm{a}+b_\mathrm{w} + \log_2 {nk^2}$. The complexity of a single output calculation consists therefore of $nk^2$ $b_\mathrm{a}$-wide $\times$ $b_\mathrm{w}$-wide multiplications and about the same amount of $b_\mathrm{o}$-wide additions. This yields the total layer complexity of
$$
	\mathrm{BOPs} \, \approx \, mnk^2( b_\mathrm{a} b_\mathrm{w} + b_\mathrm{a}+b_\mathrm{w} + \log_2 {nk^2}).
$$
Note that the reduction of the weight and activation bitwidth decreases the number of BOPs as long as the factor $b_\mathrm{a} b_\mathrm{w}$ dominates the factor $\log_2 nk^2$. Since the latter factor depends only on the layer topology, this point of diminishing return is network architecture-dependent.

\begin{table}[t]
	\caption{UNIQ accuracy on CIFAR-10 for different bitwidth.}
	\label{tab_cifar_dif_bitwidth}
	\begin{center}
		\begin{small}
			\begin{tabular}{cc|ccc}
				\toprule
				 &    & \multicolumn{3}{c}{\bf{Activation bits}}                 \\
				 &    & 4                                        & 8     & 32    \\
				\midrule
				\multirow{3}{*}{\rotatebox{90}{\makecell{\bf{Weight}             \\\bf{bits}}}}
				 & 2  & 88.1                                     & 90.88 & 89.14 \\
				 & 4  & 89.5                                     & 91.5  & 89.70 \\
				 & 32 & 88.52                                    & 91.32 & 92.00 \\
				\bottomrule
			\end{tabular}
		\end{small}
	\end{center}
\end{table}
\begin{table}[t]
	\caption{UNIQ with different quantization methods (ResNet-18 top-1 accuracy on CIFAR-10, 3-bit weights)}
	\label{other_quant_comp}
	\begin{center}
		\begin{small}
			\begin{tabular}{lcc}
				\toprule
				{\bf Quantization method} & {\bf Accuracy} & {\bf Training time [h]} \\
				\midrule
				Baseline (unquantized)    & 92.00          & 1.42                    \\ 
				$k$-quantile              & 91.30          & 2.28                    \\
				$k$-means                 & 85.80          & 5.37                    \\
				Uniform                   & 84.93          & 5.37                    \\
				\bottomrule
			\end{tabular}
		\end{small}
	\end{center}
	\vspace{-0.3cm}
\end{table}

Another factor that must be incorporated into the BOPs calculation is the cost of fetching the parameters from an an external memory. Two assumptions are made in the approximation of this cost: firstly, we assume that each parameter is only fetched once from an external memory; secondly, the cost of fetching a $b$-bit parameter is assumed to be $b$ BOPs. Given a neural network with $n$ parameters all represented in $b$ bits, the memory access cost is simply $nb$.

Figure \ref{fig_accuracy_vs_bops} and Table \ref{tab_resnet_imagenet_comparison} display the performance-complexity tradeoffs of various neural networks trained using UNIQ and other methods to different levels of weight and activation quantization.
Notice that since we quantize the first and last layers, our ResNet-34 network has better computational complexity and better accuracy compared to all competing ResNet-18 networks. The same holds for our ResNet-50 compared to all competing ResNet-34.

\subsection{Ablation study}
\label{quantization_analisys}

\paragraph{ Accuracy vs.\ quantization level.} We tested the effect of training 
ResNet-18 on CIFAR-10 with UNIQ for various levels of weight and activation quantization. Table \ref{tab_cifar_dif_bitwidth} reports the results. We observed that for such a small dataset the quantization of activations and weights helps avoid over-fitting and the results for quantized model come very close to those with full precision.

\paragraph{Comparison of different quantizers.} In the following experiment, different quantizers were compared within the uniform noise injection scheme.

The bins of the uniform quantizer were allocated evenly in the range $\left[-3\sigma,3\sigma\right]$ with $\sigma$ denoting the standard deviation of the parameters. For both the $k$-quantile and the $k$-means quantizers, normal distribution of the weights was assumed and the normal cumulative distribution and quantile functions were used for the uniformization and deuniformization of the quantized parameter.
The $k$-means and uniform quantizers used  a pre-calculated set of thresholds translated to the uniformized domain. Since the resulting bins in the uniformized domain had different widths, the level of noise was different in each bin. This required an additional step of finding the bin index for each parameter approximately doubling the training time.

The three quantizers were evaluated in a ResNet-18 network trained on the CIFAR-10 dataset with weights quantized to $3$ bits ($k=8$) and activations computed in full precision ($32$ bit).
Table \ref{other_quant_comp} reports the obtained top-1 accuracy. $k$-quantile quantization outperforms other quantization methods and is only slightly inferior to the full-precision baseline. In terms of training time, the $k$-quantile quantizer requires about $60\%$ more time to train for $k=8$; this is compared to around $280\%$ increase in training time required for the $k$-means quantizer. In addition, $k$-quantile training time is independent on the number of quantization bins as the noise distribution is same for every bin while the other methods require separate processing of each bin, increasing the training time for higher bitwidths.

\paragraph{Training from scratch vs.\ fine-tuning.}
Both training from scratch (that is, from random initialization) and fine-tuning have their advantages and disadvantages. Training from scratch takes more time but requires a single training phase with no extra training epochs, at the end of which a quantized model is obtained. Fine-tuning, on the other hand, is useful when a pre-trained full-precision model is already available; it can then be quantized with a short re-training.

Table \ref{tab_scratch_vs_finetune} compares the accuracy achieved in the two regimes on a narrow version of ResNet-18 trained on CIFAR-10 and 100. $5$-bit quantization of weights only and $5$-bit quantization of both weights and activations were compared.
We found that both regimes work equally well, reaching accuracy close to the full-precision baseline.

\begin{table}[htb]
	\caption{Top-1 accuracy (in percent) on CIFAR-10 and 100 of a narrow version on ResNet-18 trained with UNIQ from random initialization vs.\ fine-tuning a full-precision model. Number of bits is reported as (weights,activations). }
	\label{tab_scratch_vs_finetune}
	\begin{center}
		\begin{small}
			\begin{tabular}{llccccc}
				\toprule
				\bf{Dataset}               & \bf{Bits} & \bf{Full training} & \bf{Fine-tuning} & \bf{Baseline}         \\
				\midrule
				\multirow{2}{*}{CIFAR-10}  & 5,32      & 93.8               & 90.9             & \multirow{2}{*}{92.0} \\
				                           & 5,5       & 91.56              & 91.21            & \medskip              \\
				\multirow{2}{*}{CIFAR-100} & 5,32      & 66.54              & 65.73            & \multirow{2}{*}{66.3} \\
				                           & 5,5       & 65.29              & 65.05            &                       \\
				\bottomrule
			\end{tabular}
		\end{small}
	\end{center}
\end{table}

\paragraph{Accuracy vs.\ number of quantization stages.}
\label{gradual_quatization}

We found that injecting noise to all layers simultaneously does not perform well for deeper networks. As described in Section \ref{gradual_quantization_method}, we suggest splitting the training into $N$ stages, such that at each stage the noise is injected only into a subset of layers.

To determine the optimal number of stages, we fine-tuned ResNet-18 on CIFAR-10 with a fixed $18$ epoch budget. Bit width was set to $4$ for both the weights and the activations.

\begin{figure}[ht]
	\begin{center}
		\centerline{\includegraphics[width=\linewidth]{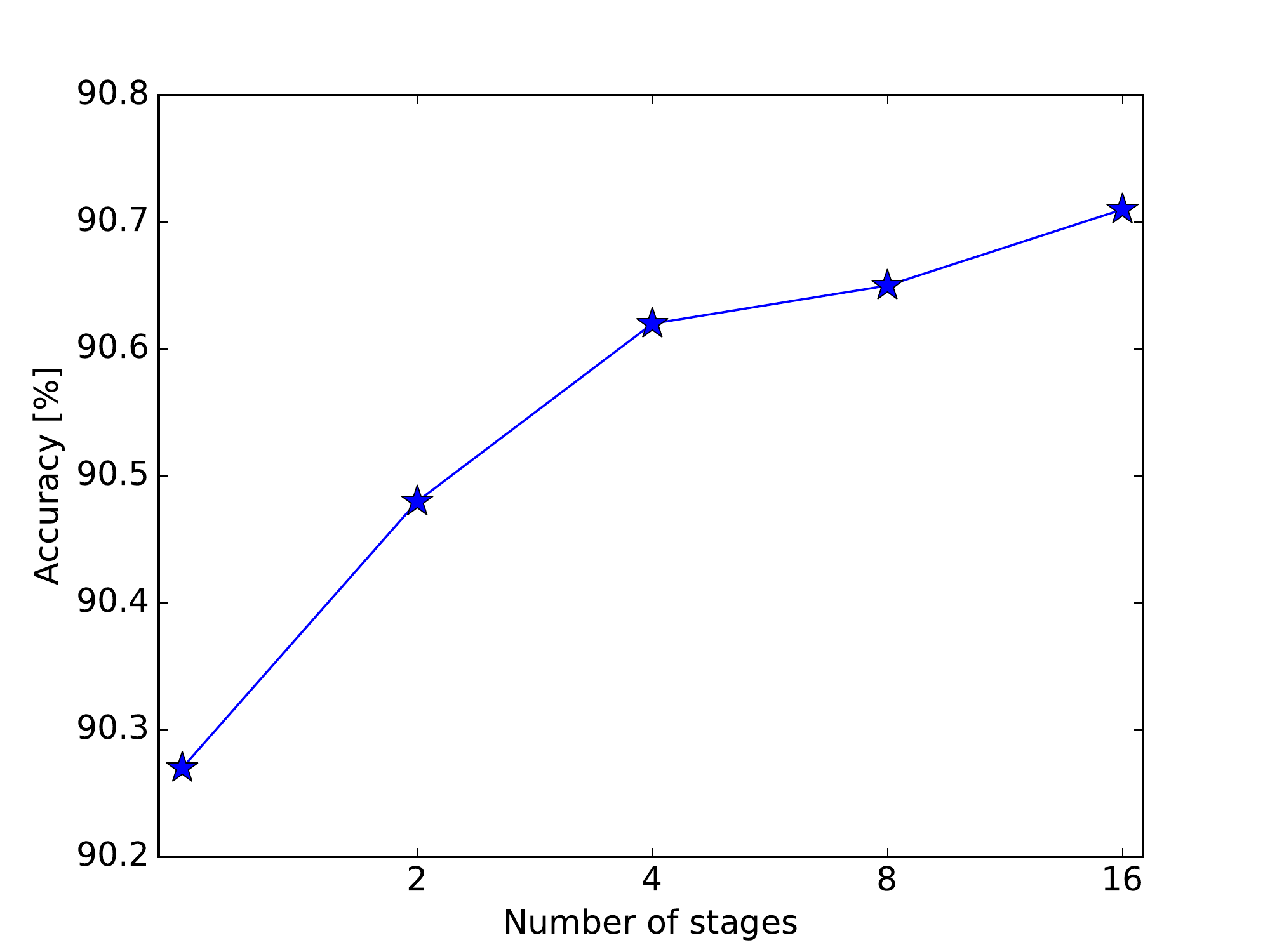}}
		\caption{Classification accuracy on CIFAR-10 of the ResNet-18 architecture quantized using UNIQ with different number of quantization stages during training. }
		\label{gradual_plot}
	\end{center}
\end{figure}

Figure \ref{gradual_plot} reports the classification accuracy as a function of the number of quantization stages. Based on these results, we conclude that the best strategy is injecting noise to a single layer at each stage. We follow this strategy in all ResNet-18 experiments conducted in this paper. For MobileNet, since it is deeper and includes $28$ layers, we chose to inject noise to $2$ consecutive layers at every stage.

\begin{figure*}[h!]
	\begin{center}
		\centerline{\includegraphics[width=\linewidth]{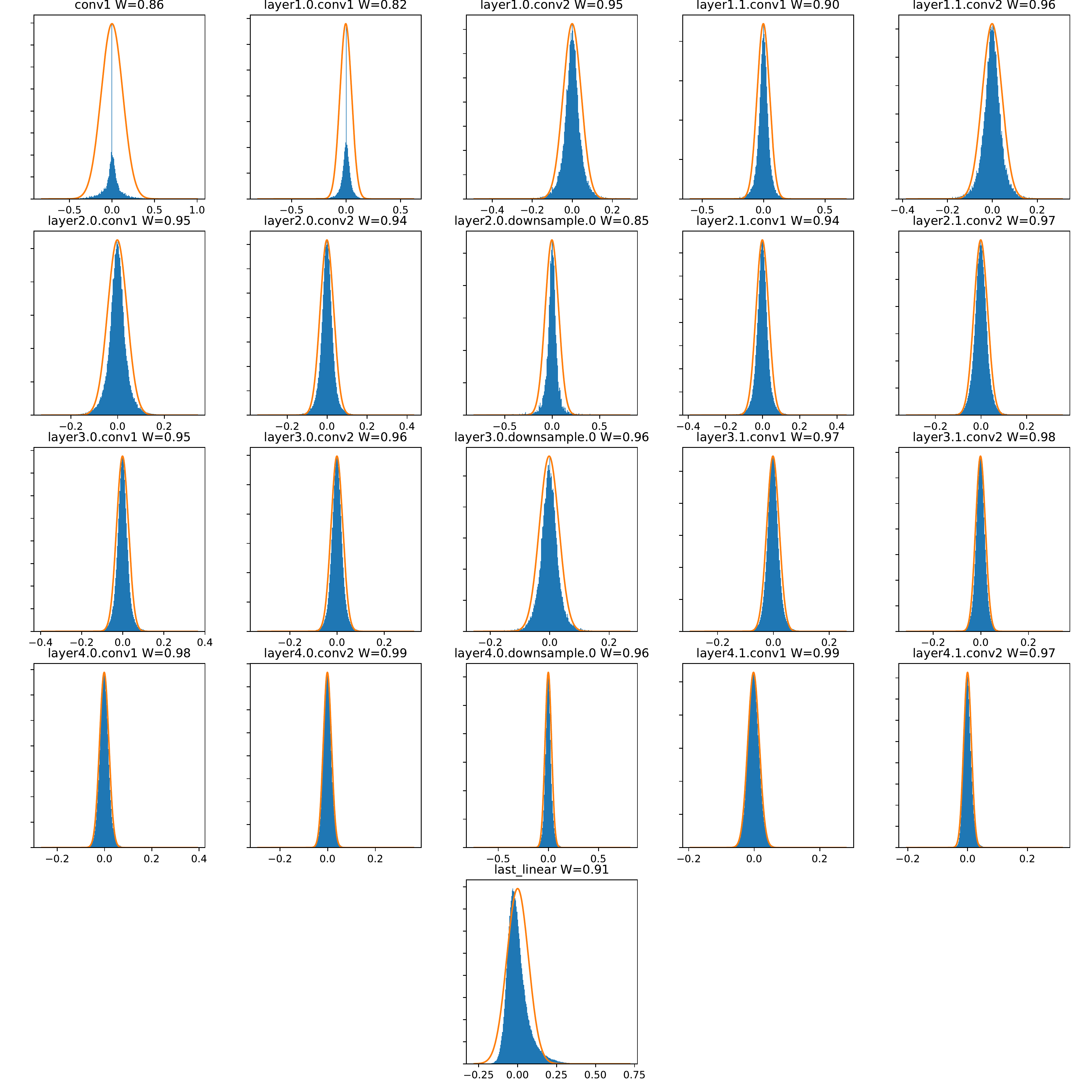}}
		\caption{Distribution of weights for the layers of pretrained ResNet-18 network. For each layer Shapiro-Wilk test statistic $W$ is mentioned.}
		\label{weight_distribution}
	\end{center}
\end{figure*}

\paragraph{Weight distribution.}
We have verified that neural networks weights follow Gaussian distribution. We observed this is true for most of the layers with Shapiro-Wilk test statistic higher than $0.82$ for all of the layers, as can be seen in Figure \ref{weight_distribution}.

\section{Conclusion}

We introduced UNIQ -- a training scheme for quantized neural networks. The scheme is based on the uniformization of the distribution of the quantized parameters, injection of additive uniform noise, followed by the de-uniformization of the obtained result. The scheme is amenable to efficient training by back propagation in full precision arithmetic, and achieves maximum efficiency with the $k$-quantile (balanced) quantizer that was investigated in this paper.

To set a common basis for comparison, we have suggested measuring the complexity of each method by BOPs. In this case, we achieved improved results for quantized neural networks on ImageNet. Our solution outperforms any other quantized network in terms of accuracy in the low computational range ($<400$ GBOPs). Our MobileNet based solution achieves $66\%$ accuracy with slightly more computational resources compared to the $50\%$ achieved with extremely quantized XNOR networks \cite{rastegari2016xnor} and QNN \cite{hubara2016quantized}.

While this paper considered a setting in which all
parameters have the same bitwidth, more complicated bit allocations will be explored in following studies. The proposed scheme is not restricted to the $k$-quantile quantizer discussed in this paper, but rather applies to any quantizer. In the general case, the noise injected into each bin is uniform, but its variance changes with the bin width.



\bibliography{UNIQ_noise_quantization}
\bibliographystyle{sysml2019}

\end{document}